\documentclass[10pt, a4paper, twocolumn]{article}
 
\usepackage[utf8]{inputenc}
\usepackage[T1]{fontenc}
\usepackage[english]{babel}
 
\usepackage{helvet} 

\usepackage[varqu]{zi4}
\usepackage{microtype}
 
\usepackage{titlesec}
\titlespacing*{\section}{0pt}{1.5ex plus 1ex minus .2ex}{0.1em}
\titlespacing*{\subsection}{0pt}{1.5ex plus 1ex minus .2ex}{0.1em}
\titlespacing*{\subsubsection}{0pt}{1.5ex plus 1ex minus .2ex}{0.1em}
 
\usepackage[table]{xcolor}
\usepackage{booktabs}
\usepackage{float}
\usepackage{graphicx}
\usepackage{subcaption}
\usepackage[labelfont={bf,small}, textfont={small}]{caption}
 
\definecolor{acqua_celeste}{RGB}{0, 160, 175}
 
\usepackage{amsmath}
\usepackage{amssymb}
\usepackage{mathtools}
\usepackage{amsthm}
\usepackage{bbold}
\usepackage{pifont}
\usepackage{multirow}
\usepackage{arydshln}
\usepackage{array}
\usepackage{makecell}
\usepackage{wrapfig}
\usepackage{tabularx}
 
\definecolor{pgreen}{rgb}{0.13, 0.55, 0.13}
\definecolor{pred}{rgb}{0.8, 0.13, 0.13}
\definecolor{diversity}{HTML}{D4E1F5}
\definecolor{recognizability}{HTML}{FFE6CC}
 
\newcommand{\xmark}{\ding{55}}
\newcommand{\cmark}{\ding{51}}
 
\newcolumntype{Y}{>{\centering\arraybackslash}X}
 
\theoremstyle{plain}

\theoremstyle{definition}

\theoremstyle{remark}

\usepackage[left=1.5cm, right=1.5cm, top=1.5cm, bottom=2cm]{geometry}
\usepackage[numbers, sort&compress]{natbib}
\usepackage{hyperref}
\usepackage[capitalize,noabbrev]{cleveref}
 
\hypersetup{
    colorlinks=true,
    linkcolor=black,
    urlcolor=acqua_celeste,
    citecolor=acqua_celeste
}

\usepackage{fancyhdr}
\usepackage{authblk}
 
\title{\huge\bfseries\vspace{-1em} Inference-Time Refinement Closes the Synthetic-Real Gap in Tabular Diffusion}
 
\author[1]{Eugenio Lomurno\textsuperscript{*}\textsuperscript{†}}
\author[1]{Filippo Balzarini\textsuperscript{†}}
\author[1]{Francesco Benelle\textsuperscript{†}}
\author[1]{Francesca Pia Panaccione\textsuperscript{†}}
\author[1]{Matteo Matteucci}
 
\affil[1]{Politecnico di Milano, AIRLab, Italy}

\date{}
 
\begin{document}
 
\twocolumn[
  \begin{@twocolumnfalse}
    \maketitle
    \vspace{-3em}
 
    \begin{abstract}
        \setlength{\parindent}{0pt}
        \setlength{\parskip}{4pt}
        \itshape
        \noindent Diffusion-based generators set the current state of the art for synthetic tabular data, deployed downstream wherever direct access to real records is restricted.
        These methods approach but rarely exceed real-data utility on downstream tasks, and closing this synthetic--real performance gap has so far been pursued exclusively at training time, via architectural advances, scaling, and retraining of monolithic generators.
        The inference-time alternative, i.e., refining the outputs of a pre-trained backbone with parameters left untouched, has remained largely unexplored for tabular synthesis.
        We introduce \textbf{TARDIS} (Tabular generation through Refinement, Distillation, and Inference-time Sampling), an inference-time refinement framework that operates on a frozen pre-trained backbone, configured per dataset by a Tree-structured Parzen Estimator search over score-level guidance during reverse diffusion, with each trial's objective set by an inner grid search over post-hoc sample selectors and an optional soft-label distillation step.
        The search space encodes a single mathematical pattern we name \textit{Bidirectional Chamfer Refinement} (BCR): the symmetric Chamfer functional between synthetic and real samples is minimized both continuously, via a score-level gradient during reverse diffusion, and discretely, via batch-ranking post-generation.
        On the majority of datasets the search selects BCR-aligned configurations over alternatives encoded in the search space, evidence both for BCR as the dominant refinement pattern and for TARDIS's per-dataset search as a procedure that recovers this pattern.
        Across 15 binary, multiclass, and regression benchmarks TARDIS achieves a median \ensuremath{+8.6\%} downstream-task improvement over models trained on real data (95\% CI \ensuremath{[+3.3, +16.4]}, Wilcoxon \ensuremath{p=0.016}, 11/15 strict wins) and improves over the underlying TabDiff backbone on all 15 datasets (mean \ensuremath{+12.9\%}, \ensuremath{p<10^{-4}}), matching the backbone on manifold fidelity, diversity, and sample-level privacy.
        The synthetic--real gap is therefore not primarily a training-time problem: on the studied corpus, inference-time refinement of a pre-trained tabular diffusion backbone reaches and exceeds real-data utility in 1 to 80 minutes on a single consumer-grade GPU.
    \end{abstract}
 
    \vspace{0.5em}
    \noindent\textbf{Keywords:} Tabular Diffusion \ensuremath{\cdot} Inference-Time Refinement \ensuremath{\cdot} Chamfer Guidance \ensuremath{\cdot} Synthetic Data
 
    \vspace{1em}
    \hrule height 1pt
    \vspace{2em}
  \end{@twocolumnfalse}
]
 
{
  \let\thefootnote\relax
  \footnotetext{\textsuperscript{*}\,Corresponding author: \texttt{eugenio.lomurno@polimi.it}}
  \footnotetext{\textsuperscript{†}\,These authors contributed equally to this work.}
}

\section{Introduction}
\label{sec:introduction}
 
Mixed-type tabular records remain the dominant input format of applied machine learning. When direct access to such records is restricted, synthetic generation enables downstream training~\cite{gdpr2018general,davila2025navigating}, and diffusion-based generators currently set the state of the art~\cite{stoian2025survey}. Despite steady architectural progress, downstream learners trained on the output of these generators approach but rarely match the utility of equivalent learners trained on real data, leaving a persistent \textit{synthetic--real gap}.
 
Existing approaches to closing this gap follow a single paradigm: improve the generator. Each new architecture, e.g., TabDiff~\cite{shi2024tabdiff} with its per-feature noise schedules, adds complexity but yields only marginal gains on the headline downstream metric, leaving real-data utility unmatched on most standard benchmarks. The implicit assumption underlying this trajectory is that the gap must be closed at training time, with the generator as the only target of intervention.

\begin{figure*}[t]
    \centering
    \includegraphics[width=0.95\textwidth]{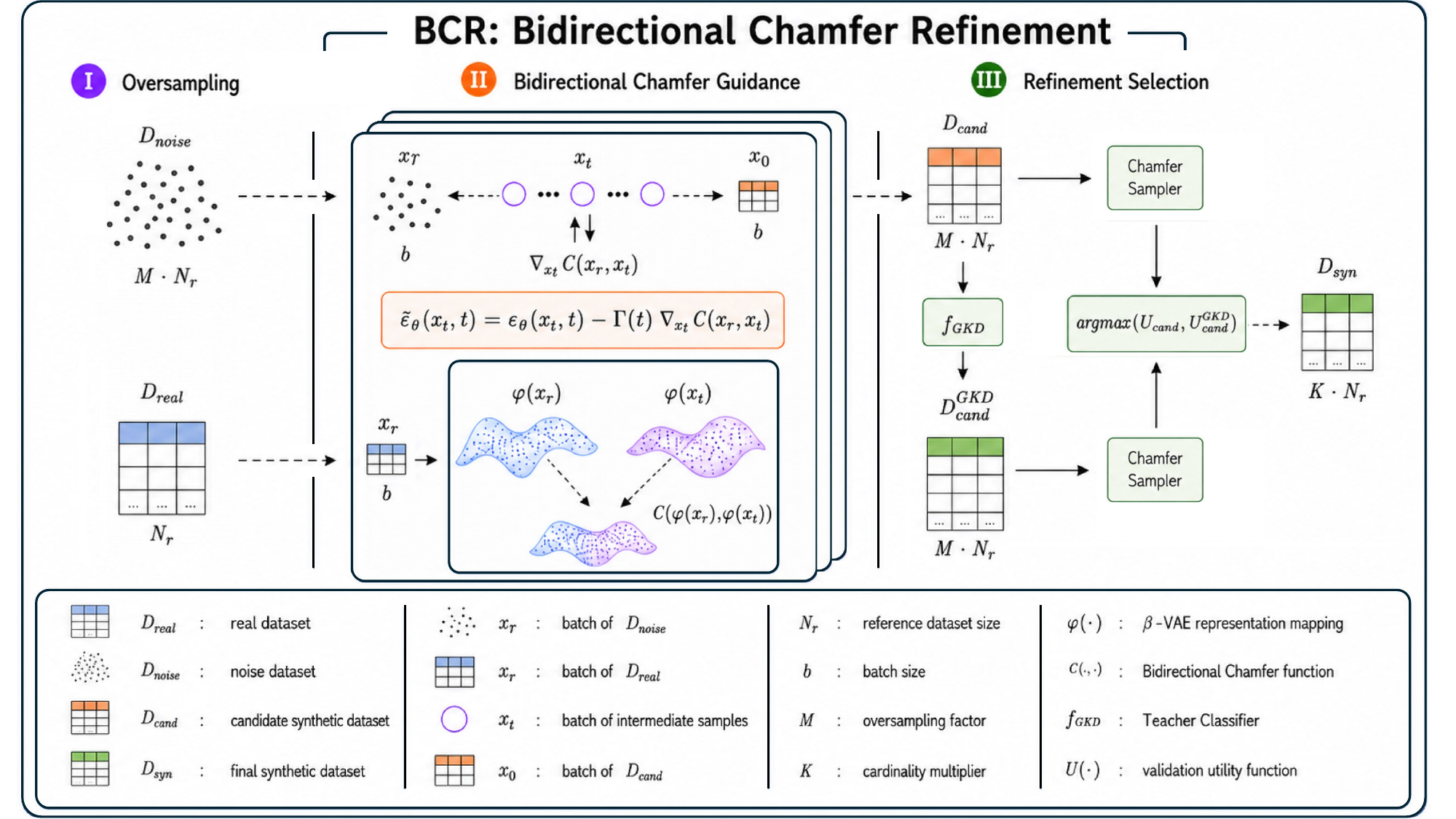}
    \caption{\textbf{TARDIS pipeline.} Stage~I draws an oversampled noise pool $D_{\mathrm{noise}}$ of cardinality $M \cdot N_r$ from the latent space $\mathcal{Z}$. Stage~II denoises $D_{\mathrm{noise}}$ via reverse diffusion, perturbing the score $\varepsilon_\theta(x_t,t)$ with the gradient of the bidirectional Chamfer functional $\mathcal{C}$ between the current candidate batch $x_t$ and a real reference batch $x_r$, projected through a representation map $\varphi$; this produces the candidate pool $D_{\mathrm{cand}}$. Stage~III selects $D_{\mathrm{syn}}$ of cardinality $K \cdot N_r$ from $D_{\mathrm{cand}}$ via a Chamfer Sampler, optionally preceded by Generative Knowledge Distillation $f_{\mathrm{GKD}}$ that relabels candidates with soft targets; the final variant is chosen by argmax over the validation utility $\mathcal{U}$. The figure illustrates the BCR-headline configuration with $\beta$-VAE mapping and Chamfer Sampler. The full TARDIS framework generalizes Stage~II to also support disabled-guidance and identity-mapping variants, and Stage~III to a grid over five samplers (Section~\ref{sec:method-stage3}, Appendix~\ref{app:samplers}).}
    \label{fig:pipeline}
\end{figure*}
 
An alternative paradigm has emerged in image synthesis, where inference-time guidance methods modify the sampling process of a pre-trained diffusion model at no additional training cost. Classifier-free guidance~\cite{ho2022classifier} steers sampling toward class-conditional or prompt-conditional outputs by interpolating score predictions; Chamfer Guidance~\cite{dall2025increasing} steers it toward arbitrary user-specified reference batches by perturbing the score with the gradient of a bidirectional Chamfer functional. Direct transfer of this paradigm to tabular synthesis is non-trivial, however, due to (i) mixed-type feature spaces that lack a canonical differentiable distance, (ii) the absence of a standardized representation space analogous to CLIP for images, and (iii) the dual fidelity-utility objective specific to the train-on-synthetic-test-on-real (TSTR) setting.
 
At its core, our approach generates many candidate samples from the pre-trained backbone and retains those that best match the real-data manifold, both by steering generation toward it and by filtering the output pool afterward. We call this framework \textbf{TARDIS} (Tabular generation through Refinement, Distillation, and Inference-time Sampling), an inference-time refinement framework that operates on a frozen pre-trained tabular diffusion backbone. TARDIS is grounded in the \textit{Bidirectional Chamfer Refinement} (BCR) principle: a continuous score-level perturbation during reverse diffusion and a discrete post-generation subsampling step both minimize the same symmetric Chamfer functional between candidates and a real reference batch, with the continuous step concentrating probability mass near the real-data manifold and the discrete step pruning residual off-manifold candidates. BCR addresses the three obstacles above in unified form: it handles (i) and (ii) through a representation map $\varphi$ instantiated per dataset either as the identity on $z$-score-normalized features or as the encoder of a tabular $\beta$-VAE, providing a differentiable representation that absorbs mixed-type heterogeneity without requiring a canonical metric; and (iii) by balancing fidelity (synthetic-to-real) and coverage (real-to-synthetic) terms in a single bidirectional functional, coupling the TSTR utility objective directly to manifold-distance minimization.
 
The framework operationalizes BCR through three stages (Figure~\ref{fig:pipeline}): Stage~I oversamples a noise pool, Stage~II applies the score-level Chamfer perturbation during reverse diffusion, and Stage~III ranks the resulting candidate pool under the same functional, optionally after a soft-label distillation step~\cite{hinton2015distilling,lomurno2025synthetic}. Stage~II hyperparameters are tuned per dataset by a Tree-structured Parzen Estimator search~\cite{bergstra2013making}; at each trial Stage~III runs a grid over selectors and the distillation toggle, returning its best validation utility as the trial's objective.
 
Empirically, across 15 heterogeneous tabular benchmarks (binary classification, multiclass classification, regression), TARDIS surpasses real-data utility on 11 of 15 datasets at a median $\mathbf{+8.6\%}$ improvement on the headline downstream metric, improves over the underlying TabDiff backbone on all 15, and matches the backbone on manifold fidelity, diversity, and sample-level privacy. The synthetic--real gap is therefore, on this corpus, an inference-time rather than a training-time problem.
 
TARDIS is, to the best of our knowledge, the first inference-time refinement framework for tabular diffusion; BCR is the unifying principle coupling its continuous and discrete steps under a single Chamfer functional; and our empirical results show that inference-time refinement suffices to close the synthetic--real gap without retraining.
 
\section{Related Work}
\label{sec:related}
 
We review prior work along three axes that intersect in TARDIS, i.e., tabular generative modelling, inference-time guidance for diffusion, and post-hoc curation of generated samples.
 
\textbf{Tabular generative models.}
Existing approaches treat the synthetic--real utility gap as a problem of the generator architecture. GAN-based methods, e.g., CTGAN~\cite{xu2019modeling} and CTAB-GAN+~\cite{zhao2024ctab}, address mixed-type generation through conditional sampling and mode-specific normalization; VAE and autoregressive families, with TVAE~\cite{xu2019modeling} and TabularARGN~\cite{sidorenko2025tabularargn}, factorize the joint via variational decoders or per-feature autoregressive heads; diffusion-based formulations, with TabDDPM~\cite{kotelnikov2023tabddpm}, STaSy~\cite{kim2022stasy}, CoDi~\cite{lee2023codi}, TabSyn~\cite{zhang2023mixed}, and TabDiff~\cite{shi2024tabdiff}, set the current state of the art through increasingly elaborate score-based and latent-space designs. Each step advances the architecture, but none operates on the output distribution of an already-trained backbone. TARDIS departs from this trajectory by refining that output distribution at inference time, with the backbone left frozen.
 
\textbf{Inference-time guidance for diffusion.}
Inference-time guidance steers a pre-trained diffusion model toward task-specific or user-specified outputs without retraining. Building on standard score-based diffusion~\cite{ho2020denoising}, classifier-free guidance~\cite{ho2022classifier} interpolates between conditional and unconditional score predictions for class- and prompt-conditional image synthesis; Chamfer Guidance~\cite{dall2025increasing} extends this to reference-based steering by perturbing the score with the gradient of a bidirectional Chamfer functional between candidates and a reference set in image space. Tabular counterparts are absent in the literature. TARDIS adapts the Chamfer construction to the tabular setting through a representation map $\varphi$ instantiated per dataset (identity on $z$-score-normalized features or the encoder of a $\beta$-VAE), and adds a discrete post-hoc selection stage that Chamfer Guidance does not include, unifying both levels under the same functional.
 
\textbf{Synthetic-data filtering and post-hoc refinement.}
Post-hoc curation of generated samples emerged in image synthesis with the Chamfer Sampler of~\cite{dall2025increasing}, which ranks an oversampled candidate pool by bidirectional manifold proximity to a real reference set. In tabular synthesis the closest precedent is ISSOSYNTH~\cite{garber2025iterative}, which oversamples a pre-trained generator and selects a subset minimizing a binned $L_1$ distance over low-order marginals via discrete iterative-proportional-fitting. Soft-label distillation has been used in tabular and image synthesis to inject teacher knowledge into synthetic training sets~\cite{hinton2015distilling,lomurno2025federated}; feature-space oversampling for class imbalance, e.g., SMOTE~\cite{chawla2002smote}, operates without a learned metric or generative pool. TARDIS reframes the oversample-and-select pattern of ISSOSYNTH as one of two coupled operational levels, pairing the discrete selection step with continuous score-level guidance during reverse diffusion, replacing binned-marginal distances with a continuous metric in $\varphi$-space, and unifying both levels under a single Chamfer functional.
 
Across these three clusters, no prior work occupies the intersection TARDIS targets: \textit{post-pretraining} refinement of a tabular diffusion backbone, with continuous and discrete steps unified under a single Chamfer functional. The closest precedents are Chamfer Guidance~\cite{dall2025increasing} in the image domain (continuous guidance only) and ISSOSYNTH~\cite{garber2025iterative} in tabular synthesis (discrete selection only); TARDIS combines and extends both.
 
\section{Methodology}
\label{sec:method}
 
We introduce \textbf{TARDIS} (Tabular generation through Refinement, Distillation, and Inference-time Sampling), a three-stage inference-time refinement framework operating on a frozen pre-trained tabular diffusion backbone $G_\theta : \mathcal{Z} \to \mathcal{X} \times \mathcal{Y}$, where $\mathcal{Z}$ is the latent noise space, $\mathcal{X} = \mathcal{X}_{\mathrm{num}} \times \mathcal{X}_{\mathrm{cat}}$ the mixed-type feature space, and $\mathcal{Y}$ the target space ($\mathbb{R}$ for regression, finite for classification). Given a real dataset $D_{\mathrm{real}} \subset \mathcal{X} \times \mathcal{Y}$ of cardinality $N_r$ and a held-out test set $D_{\mathrm{test}}$, TARDIS produces a synthetic dataset $D_{\mathrm{syn}}$ of cardinality $K \cdot N_r$ targeting the train-on-synthetic-test-on-real (TSTR) utility
\begin{equation}
\label{eq:tstr-objective}
\mathcal{U}_{\mathrm{syn}} = \psi\!\big(A(D_{\mathrm{syn}}), D_{\mathrm{test}}\big),
\end{equation}
\begin{equation}
\mathcal{U}_{\mathrm{real}} = \psi\!\big(A(D_{\mathrm{real}}), D_{\mathrm{test}}\big),
\end{equation}
where $A$ is a downstream learner and $\psi$ a task-appropriate metric (AUROC, RMSE). The synthetic--real gap is the empirical regularity $\mathcal{U}_{\mathrm{syn}} < \mathcal{U}_{\mathrm{real}}$ across mainstream tabular generators; TARDIS aims to close, and where possible reverse, this gap at inference time.
 
The framework comprises three stages (Figure~\ref{fig:pipeline}). Stage~I draws an oversampled noise pool $D_{\mathrm{noise}} \subset \mathcal{Z}$ of cardinality $M \cdot N_r$. Stage~II denoises $D_{\mathrm{noise}}$ through the reverse-diffusion process of $G_\theta$, optionally perturbing the predicted score at each step with the gradient of a bidirectional Chamfer functional against $D_{\mathrm{real}}$, yielding the candidate pool $D_{\mathrm{cand}}$ of cardinality $M \cdot N_r$. Stage~III selects $D_{\mathrm{syn}}$ from $D_{\mathrm{cand}}$ via post-hoc batch ranking under the same functional, optionally preceded by a soft-label distillation step. Stages~II and~III thus minimize the same Chamfer functional $\mathcal{C}$ at two complementary operational levels: continuously, via score-level gradient descent during sampling, and discretely, via batch ranking after sampling. We refer to this joint minimization as \textit{Bidirectional Chamfer Refinement} (BCR). Stage~II hyperparameters are tuned per dataset by a Tree-structured Parzen Estimator (TPE) search~\cite{bergstra2013making}; at each trial, Stage~III performs an inner grid search over its selector and distillation toggle, and the maximum validation utility across that grid is returned as the trial's objective.
 
\subsection{Stage I: Oversampling}
\label{sec:method-stage1}
 
Stage~I draws $M \cdot N_r$ samples from the latent prior of $\mathcal{Z}$, forming the noise pool $D_{\mathrm{noise}}$. The oversampling factor $M$ controls the redundancy available to Stage~III: each retained sample is selected from $M / K$ candidates. We fix $M = 50$ across all benchmarks. BCR requires $M > K$ so that the Stage~III refinement operates on a non-trivial selection set; the saturation of utility as $K \to M$ is characterized in Appendix~\ref{app:bcr-principle}.
 
\subsection{Stage II: Bidirectional Chamfer Guidance}
\label{sec:method-stage2}
 
Stage~II denoises $D_{\mathrm{noise}}$ through the reverse-diffusion process of $G_\theta$. During reverse diffusion the predicted score is optionally perturbed by the gradient of a bidirectional Chamfer functional against $D_{\mathrm{real}}$, steering candidates toward the real-data manifold. The resulting candidate pool $D_{\mathrm{cand}}$ has cardinality $M \cdot N_r$. The TPE search introduced above selects all Stage~II hyperparameters jointly per dataset; the full search space is reported in Appendix~\ref{app:configurations}, Table~\ref{tab:tpe-search-space}.
 
\textbf{Bidirectional Chamfer functional.}
Let $\varphi : \mathcal{X} \to \mathbb{R}^d$, with $d\in\mathbb{N}^+$, be a representation map and $A, B \subset \mathcal{X}$ two finite sample sets. We adopt the symmetric Chamfer functional~\cite{dall2025increasing},
\begin{align}
\label{eq:cc}
\mathcal{C}(A, B) \;=\;& \frac{1}{|A|} \sum_{a \in A} \min_{b \in B} \big\| \varphi(a) - \varphi(b) \big\|_2 \nonumber \\
&+\; \frac{1}{|B|} \sum_{b \in B} \min_{a \in A} \big\| \varphi(a) - \varphi(b) \big\|_2 \,,
\end{align}
which decomposes into a fidelity term penalizing deviations from $A$ to its nearest match in $B$ (first sum) and a coverage term penalizing the symmetric direction (second sum). $\mathcal{C}$ is symmetric, non-negative, and zero if and only if $\varphi(A) = \varphi(B)$ as multisets (Appendix~\ref{app:bcr-principle}).
 
\textbf{Guidance variants.}
The TPE search exposes guidance as a binary toggle. When disabled, the score perturbation of Equation~\ref{eq:score} is dropped and Stage~II reduces to standard reverse diffusion. When enabled, the representation map $\varphi$ admits two mutually exclusive forms, also exposed to the TPE search: (i) \textit{identity mapping}, with $\varphi$ the identity on $z$-score-normalized features and the Chamfer gradient computed in the data space; (ii) \textit{$\beta$-VAE mapping}, with $\varphi$ the encoder of a tabular $\beta$-VAE~\cite{kingma2013auto,burgess2018understanding} learned per dataset on $D_{\mathrm{real}}$ and the Chamfer gradient computed in its latent space. Figure~\ref{fig:pipeline} illustrates the $\beta$-VAE-mapping variant.
 
\textbf{$\beta$-VAE encoder.}
The encoder is trained on $D_{\mathrm{real}}$ under the standard $\beta$-VAE evidence lower bound,
\begin{equation}
\label{eq:vae}
\mathcal{L}_{\mathrm{VAE}}(x) \;=\; \mathcal{L}_{\mathrm{rec}}(x, \hat{x}) \;+\; \beta(t) \cdot D_{\mathrm{KL}}\!\big( q_\varphi(z \mid x) \,\|\, p(z) \big) \,,
\end{equation}
with prior $p(z) = \mathcal{N}(0, I)$, KL annealing $\beta(t)$ from one of five schedule families, and a composite reconstruction loss combining mean squared error on numerical features with cross-entropy on categorical groups, averaged within group to balance the gradient contribution across feature types. The latent dimension $d \in \{4, 8, 16, 32\}$ is selected by the TPE search; full schedule families and training hyperparameters are deferred to Appendix~\ref{app:schedules}.
 
\textbf{Score perturbation.}
At each guidance step $t$, the perturbed score is
\begin{equation}
\label{eq:score}
\tilde{\varepsilon}_\theta(x_t, t) \;=\; \varepsilon_\theta(x_t, t) \;-\; \Gamma(t) \cdot \nabla_{x_t} \mathcal{C}\big(x_t, x_r\big) \,,
\end{equation}
where $\varepsilon_\theta$ is the score predicted by the backbone, $\Gamma(t) \geq 0$ a time-dependent scaling factor, and $x_r \subset D_{\mathrm{real}}$ a reference batch sampled at the start of each generation and held fixed across diffusion steps. Guidance is applied at $t_g$ steps of the reverse trajectory rather than at every step, with $t_g$ tuned per dataset (Table~\ref{tab:tpe-search-space}); the remaining steps use the unperturbed score $\varepsilon_\theta$. Equation~\ref{eq:score} is an approximation: unlike classifier-free guidance~\cite{ho2022classifier}, it does not correspond to exact posterior sampling under a likelihood defined by $\mathcal{C}$, and $\Gamma(t)$ keeps the perturbation small relative to $\varepsilon_\theta$, acting as a soft steering signal. The schedule $\Gamma(t)$ is selected from four families (constant, linear, cosine, sine) bounded by $\gamma_{\min}, \gamma_{\max}$; the reference batch $x_r$ is either class-conditional or global.
 
\subsection{Stage III: Refinement Selection}
\label{sec:method-stage3}
 
Stage~III selects $D_{\mathrm{syn}}$ of cardinality $K \cdot N_r$ from the candidate pool $D_{\mathrm{cand}}$ produced by Stage~II. The selection is performed by an inner grid search over a binary Generative Knowledge Distillation (GKD) toggle and five sampler variants (10 combinations); for each TPE trial of Stage~II, the configuration maximizing validation utility is retained, and the resulting maximum is returned to the outer search as that trial's objective.
 
\textbf{Chamfer Sampler.}
The Chamfer Sampler is the discrete counterpart of the Stage~II score perturbation: it minimizes the same functional $\mathcal{C}$ of Equation~\ref{eq:cc}, evaluated at the pair $(B_k, D_{\mathrm{real}})$, by post-hoc batch ranking rather than gradient descent. Partitioning $D_{\mathrm{cand}}$ into batches $\{B_k\}_{k=1}^{M \cdot N_r / b}$ of size $b$,
\begin{align}
\label{eq:sampler}
\mathcal{C}(B_k, D_{\mathrm{real}}) \;=\;& \underbrace{\frac{1}{|B_k|} \sum_{x \in B_k} \min_{r \in D_{\mathrm{real}}} \|\varphi(x) - \varphi(r)\|_2}_{m_{\mathrm{sr}}(B_k):\text{ fidelity}} \nonumber \\
&+\; \underbrace{\frac{1}{|D_{\mathrm{real}}|} \sum_{r \in D_{\mathrm{real}}} \min_{x \in B_k} \|\varphi(x) - \varphi(r)\|_2}_{m_{\mathrm{rs}}(B_k):\text{ coverage}} \,,
\end{align}
batches are ranked ascending by $\mathcal{C}(B_k, D_{\mathrm{real}})$ and the top-ranked union of $K \cdot N_r$ samples constitutes $D_{\mathrm{syn}}$. The representation map $\varphi$ is inherited from Stage~II when guidance is enabled, and defaults to the identity mapping when guidance is disabled.
 
Equation~\ref{eq:sampler} and Equation~\ref{eq:score} jointly minimize the same Chamfer functional $\mathcal{C}$ on the same representation space $\varphi$ at two complementary operational levels: continuous gradient descent during reverse diffusion (Stage~II) concentrates probability mass near the real-data manifold; discrete batch ranking after diffusion (Stage~III) prunes off-manifold residuals. This joint minimization is the BCR principle; a saturation argument for the $M/K$ ratio is given in Appendix~\ref{app:bcr-principle}.
 
\textbf{Alternative samplers.}
The grid search additionally exposes four alternative samplers (Stratified, IBOSS~\cite{wang2019information}, HDBSCAN~\cite{campello2013density}, MDSampler), which differ from the Chamfer Sampler in their selection criterion: feature-space stratification, information-theoretic boundary coverage, density-based clustering, and one-directional manifold-distance ranking. Their inclusion enables the grid to recover single-stage variants when Stage~II disables guidance, in which case the Chamfer functional's bidirectional structure is no longer matched by Stage~II and a one-directional alternative may suffice. Full definitions and pseudocode are deferred to Appendix~\ref{app:samplers}.
 
\textbf{Generative Knowledge Distillation.}
GKD~\cite{hinton2015distilling,lomurno2025synthetic} replaces the hard targets in $D_{\mathrm{cand}}$ with soft-label distributions produced by a teacher classifier $f_{\mathrm{GKD}}$ trained on $D_{\mathrm{real}}$, yielding the relabeled candidate set $D_{\mathrm{cand}}^{\mathrm{GKD}}$; features are left unchanged (Appendix~\ref{app:gkd}). GKD is structurally undefined for continuous regression targets and is restricted to classification benchmarks. The grid search treats GKD as an orthogonal toggle: each sampler is evaluated on both $D_{\mathrm{cand}}$ and $D_{\mathrm{cand}}^{\mathrm{GKD}}$, and the higher-utility variant is retained.
 
\section{Experiments and Results}
\label{sec:experiments}
 
\textbf{Datasets and configurations.}
We evaluate TARDIS on 15 publicly available tabular benchmarks (UCI and Kaggle) spanning binary classification (5 datasets), multiclass classification (4), and regression (6), with structural diversity across feature counts (8 to 46), feature types, instance counts ($666$ to $53{,}940$), class balance, and domain coverage (healthcare, finance, environment, education, social science). All datasets use a 90/5/5 train/validation/test split seeded identically across models. The TPE search and inner grid of Section~\ref{sec:method} are run independently per dataset; per-dataset structural attributes and the configurations selected by the search are reported in Appendix~\ref{app:configurations} (Tables~\ref{tab:dataset_summary} and~\ref{tab:configs}, respectively).
 
\textbf{Backbone selection.}
TabDiff~\cite{shi2024tabdiff} is selected as the backbone via a controlled $1{:}1$ cardinality comparison against TabSyn~\cite{zhang2023mixed} and TabularARGN~\cite{sidorenko2025tabularargn} on the 10 benchmarks of our corpus shared with prior tabular diffusion work. Aggregating the per-dataset relative deviation from real-data utility on the headline metric, TabDiff is closest to Real with $\Delta_\% = -6.7\%$, followed by TabSyn ($-8.9\%$) and TabularARGN ($-25.7\%$). Per-dataset breakdowns and full metric coverage are deferred to Appendix~\ref{app:backbone}.
 
\textbf{Evaluation protocol.}
The downstream learner is XGBoost~\cite{chen2016xgboost} with early stopping on the validation split. We adopt the train-on-synthetic-test-on-real (TSTR) protocol, reporting AUROC for binary classification, weighted-AUROC for multiclass classification, and RMSE for regression as the primary downstream metric $\psi$. We report the per-dataset relative improvement $\Delta\%$ in $\psi$ both over the real-data baseline (sign-corrected for RMSE) and over the TabDiff backbone, and aggregate it across the 15 datasets via the median with bootstrap 95\% confidence intervals; statistical significance is assessed by Wilcoxon signed-rank paired tests, with per-dataset means as the unit of comparison. Fidelity (Precision) and diversity (Recall) are reported via manifold metrics~\cite{kynkaanniemi2019improved}; privacy via Distance to Closest Record (DCR) and Nearest-Neighbor Distance Ratio (NNDR). Per-dataset and per-metric breakdowns are deferred to Appendix~\ref{app:auxiliary-metrics}. All experiments run on a single Nvidia GTX 1080~Ti (12~GB, 2017 consumer GPU) with 5 random seeds per configuration.
 
\begin{table*}[t]
\centering
\small
\setlength{\tabcolsep}{4.5pt}
\renewcommand{\arraystretch}{1.18}
\caption{Per-dataset downstream utility under TSTR with XGBoost. Primary metric $\psi$: AUROC for BC, weighted-AUROC for MC ($\uparrow$), RMSE for R ($\downarrow$). $\Delta\%$R and $\Delta\%$T are sign-corrected relative changes vs.\ Real and TabDiff. Best per row in bold; aggregates over the 15 datasets at the bottom.}
\label{tab:results}
\begin{tabular}{@{}llcccccc@{}}
\toprule
Dataset & Task & Real & TabDiff & TARDIS$\times$1 & TARDIS & $\Delta\%$R & $\Delta\%$T \\
\midrule
\multicolumn{8}{l}{\textit{Regression --- RMSE} $\downarrow$} \\
Solar Flare        & R & 0.1025  & 0.0923  & 0.0869  & \textbf{0.0620}  & {\color{pgreen}+39.5} & {\color{pgreen}+32.8} \\
Infrared Therm.    & R & 0.0541  & 0.0585  & 0.0562  & \textbf{0.0482}  & {\color{pgreen}+10.9} & {\color{pgreen}+17.6} \\
Abalone            & R & 0.081   & 0.079   & 0.075   & \textbf{0.074}   & {\color{pgreen}+8.6}  & {\color{pgreen}+6.3} \\
Insurance          & R & 0.118   & 0.132   & 0.143   & \textbf{0.110}   & {\color{pgreen}+6.8}  & {\color{pgreen}+16.7} \\
Beijing            & R & \textbf{0.033} & 0.051   & 0.051   & 0.037   & {\color{pred}-12.1}   & {\color{pgreen}+27.5} \\
News               & R & 0.01154 & 0.01194 & 0.01000 & \textbf{0.00961} & {\color{pgreen}+16.7} & {\color{pgreen}+19.5} \\
\midrule
\multicolumn{8}{l}{\textit{Binary classification --- AUROC} $\uparrow$} \\
Diabetic Retin.    & BC & 0.6976 & 0.6671 & 0.6697 & \textbf{0.9037} & {\color{pgreen}+29.5} & {\color{pgreen}+35.5} \\
Adult              & BC & \textbf{0.927} & 0.913 & 0.922 & 0.924 & {\color{pred}-0.3} & {\color{pgreen}+1.2} \\
Default            & BC & 0.764 & 0.765 & 0.768 & \textbf{0.775} & {\color{pgreen}+1.4} & {\color{pgreen}+1.3} \\
Magic              & BC & 0.956 & 0.940 & 0.949 & \textbf{0.965} & {\color{pgreen}+0.9} & {\color{pgreen}+2.7} \\
Bank               & BC & \textbf{0.929} & 0.856 & 0.863 & 0.901 & {\color{pred}-3.0} & {\color{pgreen}+5.3} \\
\midrule
\multicolumn{8}{l}{\textit{Multiclass classification --- weighted-AUROC} $\uparrow$} \\
Student Perf.      & MC & 0.7392 & 0.7453 & 0.6796 & \textbf{0.8939} & {\color{pgreen}+20.9} & {\color{pgreen}+19.9} \\
Contraceptive      & MC & 0.6491 & 0.6838 & 0.677  & \textbf{0.7077} & {\color{pgreen}+9.0}  & {\color{pgreen}+3.5} \\
Music              & MC & 0.729  & 0.823  & 0.833  & \textbf{0.840}  & {\color{pgreen}+15.2} & {\color{pgreen}+2.1} \\
Diamonds           & MC & \textbf{0.943} & 0.930 & 0.934 & \textbf{0.943} & 0.0 & {\color{pgreen}+1.4} \\
\midrule
\multicolumn{6}{l}{\textit{Mean}}    & {\color{pgreen}+9.7}            & {\color{pgreen}+12.9} \\
\multicolumn{6}{l}{\textit{Median}}  & {\color{pgreen}\textbf{+8.6}}   & {\color{pgreen}+6.3} \\
\multicolumn{6}{l}{\textit{95\% CI}} & $[+3.3,\,+16.4]$                & $[+7.2,\,+19.2]$ \\
\multicolumn{6}{l}{$p$ (Wilcoxon)}   & $0.016$                         & $<10^{-4}$ \\
\multicolumn{6}{l}{Wins}             & 11/15                           & \textbf{15/15} \\
\bottomrule
\end{tabular}
\end{table*}
 
\begin{figure*}[t]
    \centering
    \begin{subfigure}[t]{0.49\textwidth}
        \centering
        \includegraphics[width=\linewidth]{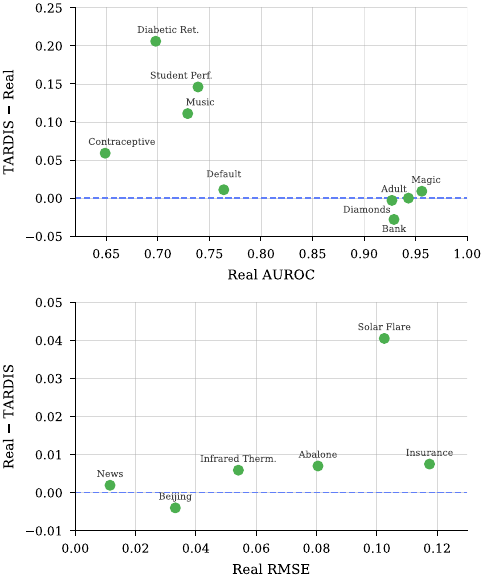}
        \caption{Per-dataset $\Delta\%$ over $\mathcal{U}_{\mathrm{real}}$ as a function of the real-data baseline, classification (top, AUROC) and regression (bottom, RMSE). Gains concentrate on datasets with low $\mathcal{U}_{\mathrm{real}}$.}
        \label{fig:headroom}
    \end{subfigure}
    \hfill
    \begin{subfigure}[t]{0.49\textwidth}
        \centering
        \includegraphics[width=\linewidth]{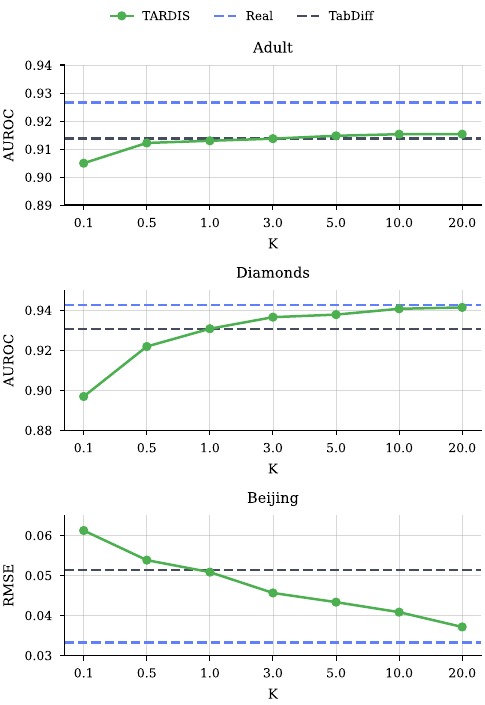}
        \caption{Cardinality saturation: sweep of $K$ with $M = 50$ on Adult, Diamonds, and Beijing, the three benchmarks where TARDIS yields the smallest gain over Real (Table~\ref{tab:results}). TARDIS still overtakes TabDiff near $K \approx 1.0$ on all three.}
        \label{fig:bcr_signatures}
    \end{subfigure}
    \caption{Empirical signatures of TARDIS performance: utility headroom (left) and cardinality saturation (right).}
    \label{fig:empirical_signatures}
\end{figure*}
 
\subsection{Main results}
\label{sec:results-headline}
 
\textbf{Aggregate downstream utility.}
Across the 15 benchmarks (Table~\ref{tab:results}), TARDIS achieves a median $\Delta\%$ improvement of $+8.6$ over $\mathcal{U}_{\mathrm{real}}$ (95\% CI $[+3.3, +16.4]$, Wilcoxon $p = 0.016$), with strict positive improvement on 11 of 15 datasets and parity on Diamonds. Against the TabDiff backbone, the improvement is uniform: $\Delta\% > 0$ on every dataset, with median $+6.3$ and mean $+12.9$ (95\% CI $[+7.2, +19.2]$, $p < 10^{-4}$). The proximity of mean and median over $\mathcal{U}_{\mathrm{real}}$ (mean $+9.7$ vs.\ median $+8.6$) indicates a tight distribution of dataset-level gains rather than an aggregate driven by outliers.
 
\textbf{Cardinality ablation.}
Setting $K = 1$ disables the oversample-and-select component of TARDIS, reducing the framework to Stage~II guidance applied at the same cardinality as $D_{\mathrm{real}}$. This restricted variant (TARDIS$\times$1, Table~\ref{tab:results}) reaches mean $\Delta\% = +1.4$ over TabDiff but does not significantly improve over $\mathcal{U}_{\mathrm{real}}$ (Wilcoxon $p > 0.05$). The full TARDIS gain therefore depends jointly on the candidate redundancy of Stage~I and on the BCR-driven discrete refinement of Stage~III; Stage~II guidance alone is insufficient to close the synthetic--real gap.
 
\textbf{Utility inverse scaling.}
Per-dataset variance in TARDIS's $\Delta\%$ over $\mathcal{U}_{\mathrm{real}}$ correlates inversely with the real-data baseline (Figure~\ref{fig:headroom}): the four datasets where the real baseline approaches the metric ceiling (Adult AUROC $0.927$, Bank $0.929$, Magic $0.956$, Diamonds $0.943$) yield $\Delta\% \in [-3.0, +0.9]$, with Bank's $-3.0$ the only negative point. Beijing represents the analogous case for regression: the target admits a near-linear structure that XGBoost on real data already exploits, leaving a residual deviation of $-12.1$. Improvement concentrates instead on datasets where $\mathcal{U}_{\mathrm{real}}$ reflects data scarcity, label noise, or class imbalance: Solar Flare ($+39.5$, $1{,}066$ instances), Diabetic Retinopathy ($+29.5$, $1{,}151$ instances), Student Performance ($+20.9$, $666$ instances), News ($+16.7$, imbalanced regression target), Music ($+15.2$, multiclass with rare classes).
 
\subsection{Empirical signatures of BCR}
\label{sec:results-bcr}
 
\textbf{Cardinality saturation.}
We sweep $K \in \{0.1, 0.5, 1, 3, 5, 10, 20\}$ with $M = 50$ fixed on three benchmarks spanning the $\mathcal{U}_{\mathrm{real}}$ regime: Adult (AUROC $0.927$), Diamonds (AUROC $0.943$), and Beijing (RMSE $0.033$). All three datasets exhibit monotone improvement in $\psi$ (Figure~\ref{fig:bcr_signatures}), with knees at $K \approx 3$ on Adult and $K \approx 5$ on Diamonds; on Beijing the curve continues to descend through $K = 20$ without plateauing within the swept range. TARDIS overtakes TabDiff near $K \approx 1.0$ on all three datasets, indicating that BCR refinement begins to add value already at real cardinality. The plateau location is ordered consistently with the strength of $\mathcal{U}_{\mathrm{real}}$: Adult saturates early below Real, Diamonds reaches parity with Real near $K = 20$, and Beijing remains below Real within the swept range.
 
\textbf{Sampler-guidance coupling.}
The TPE search and inner grid expose a structural alignment between Stage~II guidance and Stage~III sampler choice, recovered without explicit constraint (Table~\ref{tab:configs}). Of the 9 datasets where guidance is enabled, all 9 select the Chamfer Sampler in Stage~III; of the 6 datasets where guidance is disabled, 4 (Bank, Magic, Music, Abalone) select the MDSampler, and the remaining 2 (Adult, Solar Flare) select the Chamfer Sampler with GKD enabled (the only configurations with $K \leq 1$, where the candidate pool is too sparse for the MDSampler's purely fidelity-driven selection to retain coverage). The configurations therefore cluster into two coherent regimes: guided pool with bidirectional Chamfer-based selection ($9/15$), and unguided pool with one-directional manifold-distance selection ($4/15$). When Stage~II shapes the candidate pool toward the real manifold via $\nabla \mathcal{C}$, the discrete sampler that minimizes the same $\mathcal{C}$ dominates; when Stage~II is disabled and $K \gg 1$, the simpler one-directional MDSampler suffices.
 
\textbf{Low-cardinality regime.}
The configuration-level evidence in Table~\ref{tab:configs} surfaces a regime separation along $K$ that we term the \textit{low-cardinality regime} ($K \leq 1$). Across the 15 benchmarks, GKD is enabled on 2 datasets only, both with $K \leq 1$ (Adult $K{=}0.5$, Solar Flare $K{=}0.1$); on the remaining 13, all with $K \geq 8.5$, the search disables GKD. The pattern is consistent with the qualitative mechanism: at low $K$ the candidate pool is too sparse for the Chamfer Sampler to recover utility through selection alone, and the soft-label channel becomes the dominant signal; at high $K$ the pool is large enough for selection to suffice, and GKD's smoothing becomes a net negative. The analysis is restricted to classification, since GKD is structurally undefined for continuous regression targets and is consistently disabled on the 6 regression datasets in the corpus. We position GKD as a regime-specific component for hardware-constrained downstream applications, complementing the main result rather than competing with it.
 
\subsection{Fidelity, diversity, privacy, and runtime}
\label{sec:results-fpr}
 
\begin{table}[t]
\centering
\footnotesize
\setlength{\tabcolsep}{3pt}
\renewcommand{\arraystretch}{1.05}
\caption{Fidelity, diversity, and privacy proxies across 15 datasets.}
\label{tab:fidelity_privacy}
\begin{tabular}{@{}lcc@{}}
\toprule
Metric & TabDiff & TARDIS \\
\midrule
Precision $\uparrow$ & $0.947{\pm}0.073$ & $\mathbf{0.949{\pm}0.069}$ \\
Recall $\uparrow$    & $0.994{\pm}0.005$ & $\mathbf{0.995{\pm}0.005}$ \\
DCR-1 share          & $\mathbf{0.501{\pm}0.010}$ & $\mathbf{0.499{\pm}0.013}$ \\
NNDR $\uparrow$      & $0.823{\pm}0.115$ & $\mathbf{0.824{\pm}0.107}$ \\
\bottomrule
\end{tabular}
\end{table}
 
\textbf{Fidelity and diversity.}
Manifold Precision (fidelity) and Recall (diversity)~\cite{kynkaanniemi2019improved} are preserved at the TabDiff backbone level on all 15 datasets (Table~\ref{tab:fidelity_privacy}): mean Precision $0.949$ (TARDIS) vs.\ $0.947$ (TabDiff), mean Recall $0.995$ vs.\ $0.994$, with per-dataset absolute deviations bounded by $0.043$ and $0.003$ respectively. The BCR-driven refinement does not collapse the synthetic distribution onto a subset of the real manifold, the canonical failure mode of utility-aware curation.
 
\textbf{Privacy.}
DCR-1 share and NNDR are likewise preserved at the backbone level (Table~\ref{tab:fidelity_privacy}; per-dataset breakdown in Table~\ref{tab:privacy-breakdown}): TARDIS DCR-1 share $0.499 \pm 0.013$ vs.\ TabDiff $0.501 \pm 0.010$, NNDR $0.824 \pm 0.107$ vs.\ $0.823 \pm 0.115$, with no per-dataset deviation exceeding $0.04$. TARDIS does not improve privacy (which is not a claim of this paper) and the post-hoc selection mechanism does not degrade it either: although the Chamfer Sampler ranks candidates by proximity to a real reference batch, the bidirectional structure of $\mathcal{C}$ balances fidelity against coverage, preventing convergence onto training records.
 
\textbf{Runtime.}
On the Nvidia GTX 1080~Ti, TARDIS adds 0.9 to 79.4 minutes of wall-clock time on top of the trained backbone, with a median of 16.1 minutes across the 15 datasets; the largest case (Bank, $\sim\!45{,}000$ instances) runs in 79.4 minutes and the smallest (Solar Flare) in $0.9$ minutes (per-stage breakdown in Appendix~\ref{app:runtime}). Stage~II generation accounts for 47\% to 90\% of the wall-clock budget; Stage~III selection contributes 5\% to 38\%, with the highest fractions on Music and News (both $38\%$) where the candidate pool is large; GKD distillation contributes 7\% to 11\% on the two datasets where it is active. The dominant cost factor is the $M = 50$ oversampling, which is GPU-throughput-bound and shrinks proportionally on modern hardware.
 
\section{Discussion and Limitations}
\label{sec:discussion}
 
\textbf{BCR as a domain-transferable principle.}
The Chamfer functional of Equation~\ref{eq:cc} is modality-agnostic: it requires only a differentiable representation map $\varphi$ and a real-data reference batch. Its two operationalizations (continuous via $\nabla_{x_t}\mathcal{C}$ during reverse diffusion, discrete via batch ranking after sampling) extend to image, text, and joint multimodal diffusion under domain-appropriate encoders, since neither a differentiable score nor a candidate pool is tabular-specific. The configuration-level signature of Section~\ref{sec:results-bcr}, with the unconstrained search electing the joint guided-and-Chamfer regime in 9 of 15 cases, is the empirical claim BCR makes about its generality. Two further findings sharpen this picture. First, the TARDIS$\times$1 ablation (Section~\ref{sec:results-headline}) isolates Stage~II from Stage~III: continuous guidance alone yields mean $+1.4\%$ over TabDiff but does not surpass real-data utility, while the full pipeline with bidirectional discrete selection does. The discrete level thus carries the dominant share of the gap-closing effect, inverting the image-domain emphasis on score-level guidance. Second, sample-level privacy is preserved as a structural property of the bidirectional functional: the coverage term penalizes distributional collapse onto $D_{\mathrm{real}}$, so ranking by proximity at Stage~III cannot drive synthetic samples onto training records (Section~\ref{sec:results-fpr}). Refinement is therefore complementary to backbone improvement, not a substitute: any competent tabular diffusion backbone admits this inference-time refinement at fixed compute overhead, with gains concentrated on the regimes where synthetic data is most operationally needed (small corpora, label noise, class imbalance).
 
\textbf{Limitations and Future Work.}
TARDIS is empirically validated on a single backbone (TabDiff); BCR is backbone-agnostic by construction, but transfer to TabSyn (latent-space diffusion) and TabDDPM (multinomial diffusion) is not empirically tested, and Stage~II's score perturbation does not apply to autoregressive (TabularARGN) or variational (TVAE) decoders. The oversampling factor $M$ is fixed at $50$, leaving the $M$-dependence of the saturation knee uncharacterized. The headline configurations use deterministic samplers, so per-seed variance reflects pool generation only, and downstream-learner transfer beyond XGBoost is not tested. Privacy is reported through DCR-1 and NNDR proxies, without formal differential-privacy claims or membership-inference evaluation. Future work targets three directions: cross-backbone and cross-modality transfer (TabSyn, TabDDPM, image-domain diffusion); an $M$-sweep enabling an adaptive recipe driven by $N_r$, manifold-coverage gap, and deployment cardinality budget; and a differentially-private variant consuming calibrated budget at both refinement levels.
 
\section{Conclusions}
\label{sec:conclusions}
 
We introduced TARDIS, an inference-time refinement framework for pre-trained tabular diffusion backbones, grounded in the \textit{Bidirectional Chamfer Refinement} (BCR) principle: a single Chamfer functional minimized continuously during reverse diffusion via score-level guidance and discretely after sampling via batch ranking, with the per-dataset search recovering this joint structure on 9 of 15 benchmarks and selecting single-stage variants on the remaining 6. Across 15 heterogeneous tabular benchmarks (binary, multiclass, regression), TARDIS achieves a median $+8.6\%$ improvement in $\psi$ over models trained on real data (Wilcoxon $p = 0.016$, $11/15$ strict wins) and improves over the TabDiff backbone on every dataset (mean $+12.9\%$, $p < 10^{-4}$), with fidelity, diversity, and privacy preserved at the backbone level and a wall-clock overhead under 80 minutes on a 2017 consumer GPU. The synthetic--real gap is therefore not primarily a training-time problem on the studied corpus, and BCR provides a domain-transferable principle for inference-time refinement of pre-trained generative models.
 
\bibliographystyle{plainnat}
\bibliography{references}

@article{kingma2013auto,
  title={Auto-encoding variational bayes},
  author={Kingma, Diederik P and Welling, Max},
  journal={arXiv preprint arXiv:1312.6114},
  year={2013}
}

@article{shi2024tabdiff,
  title={Tabdiff: a mixed-type diffusion model for tabular data generation},
  author={Shi, Juntong and Xu, Minkai and Hua, Harper and Zhang, Hengrui and Ermon, Stefano and Leskovec, Jure},
  journal={arXiv preprint arXiv:2410.20626},
  year={2024}
}

@article{zhang2023mixed,
  title={Mixed-type tabular data synthesis with score-based diffusion in latent space},
  author={Zhang, Hengrui and Zhang, Jiani and Srinivasan, Balasubramaniam and Shen, Zhengyuan and Qin, Xiao and Faloutsos, Christos and Rangwala, Huzefa and Karypis, George},
  journal={arXiv preprint arXiv:2310.09656},
  year={2023}
}

@inproceedings{sidorenko2025tabularargn,
  title={TabularARGN: An Auto-Regressive Generative Network for Tabular Data Generation},
  author={Sidorenko, Andrey and Krchova, Ivona and Vieyra, Mariana Vargas and Tiwald, Paul and Scriminaci, Mario and Platzer, Michael},
  booktitle={EurIPS 2025 Workshop: AI for Tabular Data},
  year={2025}
}

@article{xu2019modeling,
  title={Modeling tabular data using conditional gan},
  author={Xu, Lei and Skoularidou, Maria and Cuesta-Infante, Alfredo and Veeramachaneni, Kalyan},
  journal={Advances in neural information processing systems},
  volume={32},
  year={2019}
}

@article{zhao2024ctab,
  title={Ctab-gan+: Enhancing tabular data synthesis},
  author={Zhao, Zilong and Kunar, Aditya and Birke, Robert and Van der Scheer, Hiek and Chen, Lydia Y},
  journal={Frontiers in big Data},
  volume={6},
  pages={1296508},
  year={2024},
  publisher={Frontiers Media SA}
}

@article{chawla2002smote,
  title={SMOTE: synthetic minority over-sampling technique},
  author={Chawla, Nitesh V and Bowyer, Kevin W and Hall, Lawrence O and Kegelmeyer, W Philip},
  journal={Journal of artificial intelligence research},
  volume={16},
  pages={321--357},
  year={2002}
}

@article{kim2022stasy,
  title={Stasy: Score-based tabular data synthesis},
  author={Kim, Jayoung and Lee, Chaejeong and Park, Noseong},
  journal={arXiv preprint arXiv:2210.04018},
  year={2022}
}

@inproceedings{kotelnikov2023tabddpm,
  title={Tabddpm: Modelling tabular data with diffusion models},
  author={Kotelnikov, Akim and Baranchuk, Dmitry and Rubachev, Ivan and Babenko, Artem},
  booktitle={International conference on machine learning},
  pages={17564--17579},
  year={2023},
  organization={PMLR}
}

@inproceedings{lee2023codi,
  title={Codi: Co-evolving contrastive diffusion models for mixed-type tabular synthesis},
  author={Lee, Chaejeong and Kim, Jayoung and Park, Noseong},
  booktitle={International Conference on Machine Learning},
  pages={18940--18956},
  year={2023},
  organization={PMLR}
}

@article{lomurno2025synthetic,
  title={Synthetic image learning: Preserving performance and preventing membership inference attacks},
  author={Lomurno, Eugenio and Matteucci, Matteo},
  journal={Pattern Recognition Letters},
  volume={190},
  pages={52--58},
  year={2025},
  publisher={Elsevier}
}

@article{lomurno2025federated,
  title={Federated Knowledge Recycling: Privacy-preserving synthetic data sharing},
  author={Lomurno, Eugenio and Matteucci, Matteo},
  journal={Pattern Recognition Letters},
  volume={190},
  pages={124-130},
  year={2025},
  publisher={Elsevier}
}

@article{wang2019information,
  title={Information-based optimal subdata selection for big data linear regression},
  author={Wang, HaiYing and Yang, Min and Stufken, John},
  journal={Journal of the American Statistical Association},
  volume={114},
  number={525},
  pages={393--405},
  year={2019},
  publisher={Taylor \& Francis}
}

@article{dall2025increasing,
  title={Increasing the utility of synthetic images through chamfer guidance},
  author={Dall'Asen, Nicola and Zhang, Xiaofeng and Hemmat, Reyhane Askari and Hall, Melissa and Verbeek, Jakob and Romero-Soriano, Adriana and Drozdzal, Michal},
  journal={arXiv preprint arXiv:2508.10631},
  year={2025}
}

@article{burgess2018understanding,
  title={Understanding disentangling in Beta-VAE},
  author={Burgess, Christopher P and Higgins, Irina and Pal, Arka and Matthey, Loic and Watters, Nick and Desjardins, Guillaume and Lerchner, Alexander},
  journal={arXiv preprint arXiv:1804.03599},
  year={2018}
}

@article{gdpr2018general,
  title={General data protection regulation (gdpr)},
  author={GDPR, EU},
  journal={Cit. on},
  pages={4},
  year={2018}
}

@article{hinton2015distilling,
  title={Distilling the knowledge in a neural network},
  author={Hinton, Geoffrey and Vinyals, Oriol and Dean, Jeff},
  journal={arXiv preprint arXiv:1503.02531},
  year={2015}
}

@article{ho2020denoising,
  title={Denoising diffusion probabilistic models},
  author={Ho, Jonathan and Jain, Ajay and Abbeel, Pieter},
  journal={Advances in neural information processing systems},
  volume={33},
  pages={6840--6851},
  year={2020}
}

@article{davila2025navigating,
  title={Navigating tabular data synthesis research understanding user needs and tool capabilities},
  author={Davila R, Maria F and Groen, Sven and Panse, Fabian and Wingerath, Wolfram},
  journal={ACM SIGMOD Record},
  volume={53},
  number={4},
  pages={18--35},
  year={2025},
  publisher={ACM New York, NY, USA}
}

@article{stoian2025survey,
  title={A survey on tabular data generation: Utility, alignment, fidelity, privacy, and beyond},
  author={Stoian, Mihaela C{\"A} and Giunchiglia, Eleonora and Lukasiewicz, Thomas},
  journal={arXiv preprint arXiv:2503.05954},
  year={2025}
}

@inproceedings{chen2016xgboost,
  title={Xgboost: A scalable tree boosting system},
  author={Chen, Tianqi and Guestrin, Carlos},
  booktitle={Proceedings of the 22nd acm sigkdd international conference on knowledge discovery and data mining},
  pages={785--794},
  year={2016}
}

@inproceedings{bergstra2013making,
  title={Making a science of model search: Hyperparameter optimization in hundreds of dimensions for vision architectures},
  author={Bergstra, James and Yamins, Daniel and Cox, David},
  booktitle={International conference on machine learning},
  pages={115--123},
  year={2013},
  organization={PMLR}
}

@article{kynkaanniemi2019improved,
  title={Improved precision and recall metric for assessing generative models},
  author={Kynk{\"a}{\"a}nniemi, Tuomas and Karras, Tero and Laine, Samuli and Lehtinen, Jaakko and Aila, Timo},
  journal={Advances in neural information processing systems},
  volume={32},
  year={2019}
}

@inproceedings{garber2025iterative,
  title={Iterative Subset Selection for High-fidelity Synthetic Tabular Data},
  author={G{"a}rber, Daniel and Demelius, Lea},
  booktitle={EurIPS 2025 Workshop: AI for Tabular Data},
  year={2025}
}

@article{ho2022classifier,
  title={Classifier-free diffusion guidance},
  author={Ho, Jonathan and Salimans, Tim},
  journal={arXiv preprint arXiv:2207.12598},
  year={2022}
}

@inproceedings{campello2013density,
  title={Density-based clustering based on hierarchical density estimates},
  author={Campello, Ricardo JGB and Moulavi, Davoud and Sander, J{\"o}rg},
  booktitle={Pacific-Asia Conference on Knowledge Discovery and Data Mining},
  year={2013}
}
 
\onecolumn
\appendix
 
\section{Bidirectional Chamfer Refinement: Properties and Saturation}
\label{app:bcr-principle}
 
This appendix collects the structural properties of the symmetric Chamfer functional $\mathcal{C}$ (Section~\ref{app:bcr-functional}), the saturation argument that motivates the empirical knee structure of Figure~\ref{fig:bcr_signatures} (Section~\ref{app:bcr-argument}), and its empirical validation on the cardinality sweep of Section~\ref{sec:results-bcr} (Section~\ref{app:bcr-empirical}).
 
\subsection{Properties of the Symmetric Chamfer Functional}
\label{app:bcr-functional}
 
Let $\varphi: \mathcal{X} \to \mathbb{R}^d$ be a representation map and $A, B \subset \mathcal{X}$ finite multisets with $\varphi$ injective on $A \cup B$. The functional $\mathcal{C}$ of Equation~\ref{eq:cc} satisfies: \textbf{(P1) non-negativity}, $\mathcal{C}(A,B) \geq 0$ (Euclidean norm under uniform weighting); \textbf{(P2) identity of indiscernibles}, $\mathcal{C}(A,B) = 0 \iff \varphi(A) = \varphi(B)$ as multisets, since fidelity and coverage terms vanish jointly only when the two sets coincide in $\varphi$-space; \textbf{(P3) symmetry}, $\mathcal{C}(A,B) = \mathcal{C}(B,A)$ by construction. P1--P3 jointly justify the use of $\mathcal{C}$ as a single optimization target: minimizing $\mathcal{C}(D_{\mathrm{syn}}, D_{\mathrm{real}})$ at the continuous (Stage~II) and discrete (Stage~III) refinement levels is a coherent objective, since any reduction at one level corresponds to monotone movement toward the same target distribution in $\varphi$-space.
 
\subsection{The BCR Saturation Argument}
\label{app:bcr-argument}
 
We argue that, under reasonable assumptions on the candidate pool and the discrete sampler, $\mathcal{U}_{\mathrm{syn}}$ exhibits a monotone-then-plateau dependence on the subsampling fraction $K$, with the knee location governed by the $M/K$ ratio. We state this as a principled argument rather than a formal theorem: we do not characterize the convergence rate as $M \to \infty$, the empirical knee depends on the intrinsic dimension of $D_{\mathrm{real}}$ in $\varphi$-space which we do not formalize, and we hold $M = 50$ fixed throughout our experiments.
 
\textbf{Argument (BCR Saturation).} Under the assumptions that (i) Stage~II concentrates the empirical density of $\varphi(D_{\mathrm{cand}})$ around the support of $\varphi(D_{\mathrm{real}})$, (ii) Stage~III selects the top $K/M$ fraction of $D_{\mathrm{cand}}$ by $\mathcal{C}$-rank, and (iii) the downstream learner is consistent in the cardinality of training data, we expect $\mathcal{U}_{\mathrm{syn}}$ to be monotonically non-decreasing in $M/K$ and to saturate at $K = K^*$ for a dataset-dependent knee $K^* \leq M$.
 
\textbf{Reasoning.} As $K$ decreases relative to $M$, the Stage~III sampler retains only the most $\mathcal{C}$-aligned candidates, sharpening $\varphi(D_{\mathrm{syn}})$ around the support of $\varphi(D_{\mathrm{real}})$ and reducing the fidelity term $m_{\mathrm{sr}}$ of Equation~\ref{eq:sampler}; as $K \to M$, the selection becomes vacuous and the discrete refinement contributes negligibly. Saturation occurs when the coverage term $m_{\mathrm{rs}}$ becomes the binding constraint, i.e., when a smaller selection fraction risks under-covering isolated regions of the real-data manifold. The knee $K^*$ is expected to scale with the effective intrinsic dimension of $D_{\mathrm{real}}$ in $\varphi$-space and with the inverse local density of real samples.
 
\subsection{Empirical Validation}
\label{app:bcr-empirical}
 
The cardinality sweep $K \in \{0.1, 0.5, 1, 3, 5, 10, 20\}$ with $M = 50$ on Adult, Diamonds, and Beijing (Figure~\ref{fig:bcr_signatures}) verifies the predicted monotone-then-plateau profile. The knees occur at $K^* \approx 3$ for Adult, $K^* \approx 5$ for Diamonds, and beyond $K = 20$ for Beijing, an ordering consistent with the intrinsic-dimension intuition (Adult mixed-type but compact; Beijing combines hourly cycles and multi-scale meteorology, yielding the highest effective dimension).
 
\section{Backbone Selection}
\label{app:backbone}
 
The choice of TabDiff~\cite{shi2024tabdiff} as the underlying backbone for TARDIS results from a controlled head-to-head comparison against TabSyn~\cite{zhang2023mixed} and TabularARGN~\cite{sidorenko2025tabularargn} on 10 of the 15 corpus benchmarks (Adult, Default, Magic, Bank, Music, Diamonds, Abalone, Insurance, Beijing, News), for which all three baselines have been independently trained at matched 1:1 cardinality. Each backbone was evaluated under the same TSTR protocol of Section~\ref{sec:experiments}, with multiple utility and fidelity metrics per dataset.
 
\textbf{Aggregate dominance.} Of the 50 metric cells, TabDiff dominates on 23, TabSyn on 14, and TabularARGN on 13, with dominance concentrated on the primary TSTR metrics (classification AUROC, regression RMSE). TabularARGN remains competitive on Bank classification, where its autoregressive factorization handles class-conditional distributions effectively, and on News regression, where the high-dimensional feature space (46 attributes) favors per-feature autoregressive heads over latent-space diffusion.
 
\textbf{Architecture and TARDIS compatibility.} TabDiff's per-feature noise schedules differentiate noising rates across features of heterogeneous scale and cardinality, avoiding the bias that a single shared schedule introduces on mixed-type inputs; its continuous reverse process and exposed score $\varepsilon_\theta$ make it directly compatible with the Stage~II score perturbation of Equation~\ref{eq:score}. TabSyn is similarly compatible via latent-space score perturbation, but loses the per-feature inductive bias. TabularARGN and variational generators (e.g., TVAE) are excluded from Stage~II since the perturbation presumes a continuous reverse process; Stage~III remains compatible with any generator producing a candidate pool, so the discrete level of BCR transfers to those generators independently.
 
\begin{table}[h]
\centering
\footnotesize
\setlength{\tabcolsep}{3pt}
\renewcommand{\arraystretch}{1.2}
\caption{Backbone comparison on 10 corpus benchmarks under matched 1:1 cardinality. Primary downstream metric reported: AUROC $\uparrow$ for binary classification, weighted-AUROC $\uparrow$ for multiclass classification, RMSE $\downarrow$ for regression. Bold indicates the best generator per column, excluding Real. The final column reports the average percentage delta with respect to Real, computed as $100(m-r)/r$ for AUROC and $100(r-m)/r$ for RMSE; higher is better.}
\label{tab:backbone-mle}
\begin{tabular}{@{}lccccccccccc@{}}
\toprule
& \multicolumn{6}{c}{Classification (AUROC $\uparrow$)} & \multicolumn{4}{c}{Regression (RMSE $\downarrow$)} & \multicolumn{1}{c}{$\Delta_\%$ vs Real $\uparrow$} \\
\cmidrule(lr){2-7}\cmidrule(lr){8-11}\cmidrule(l){12-12}
Model & Adult & Default & Magic & Bank & Music & Diamonds & Abalone & Insurance & Beijing & News & Avg. \\
\midrule
Real         & .927 & .764 & .956 & .929 & .729 & .943 & .081 & .118 & .033 & .01154 & -- \\
\midrule
TabularARGN  & \textbf{.914} & .751 & .909 & \textbf{.911} & .714 & .907 & .096 & .281 & .062 & \textbf{.01117} & $-25.72$ \\
TabSyn       & .908 & \textbf{.768} & \textbf{.942} & .896 & .724 & .929 & \textbf{.079} & \textbf{.132} & .055 & .01201 & $-8.88$ \\
TabDiff      & .913 & .765 & .940 & .856 & \textbf{.823} & \textbf{.930} & \textbf{.079} & \textbf{.132} & \textbf{.051} & .01194 & $\mathbf{-6.68}$ \\
\bottomrule
\end{tabular}
\end{table}
 
\section{Stage II Schedules and Reference Batching}
\label{app:schedules}
 
This appendix specifies the schedule families exposed to the TPE search: the guidance scaling $\Gamma(t)$ of Stage~II (Section~\ref{app:gamma}), the KL annealing weight $\beta(t)$ used during $\beta$-VAE training of $\varphi$ (Section~\ref{app:beta}), and the reference batching strategies (Section~\ref{app:batching}). Bounds $\gamma_{\min}/\gamma_{\max}$ and $\beta_{\min}/\beta_{\max}$ are selected per dataset.
 
\subsection{Guidance Schedules $\Gamma(t)$}
\label{app:gamma}
 
Let $T$ denote the total number of reverse-diffusion steps, with $t \in [0, T]$ counting from $t = T$ (pure noise) to $t = 0$ (final sample). We define four functional families for $\Gamma(t)$, each parameterized by $(\gamma_{\min}, \gamma_{\max})$ and bounded in $[\gamma_{\min}, \gamma_{\max}]$:
\begin{align}
\textbf{Constant:} \quad & \Gamma(t) = \gamma_{\max} \\
\textbf{Linear:} \quad & \Gamma(t) = \gamma_{\min} + (\gamma_{\max} - \gamma_{\min}) \cdot \frac{t}{T} \\
\textbf{Cosine:} \quad & \Gamma(t) = \gamma_{\min} + \frac{1}{2} (\gamma_{\max} - \gamma_{\min}) \left(1 + \cos\!\left(\pi \cdot \tfrac{t}{T}\right)\right) \\
\textbf{Sine:} \quad & \Gamma(t) = \gamma_{\max} - \frac{1}{2} (\gamma_{\max} - \gamma_{\min}) \left(1 + \cos\!\left(\pi \cdot \tfrac{t}{T}\right)\right)
\end{align}
 
\textbf{Behavioral characterization.} Constant applies uniform guidance across the trajectory. Linear and Sine rise monotonically from $\gamma_{\min}$ at $t=0$ to $\gamma_{\max}$ at $t=T$ (Sine with cosine curvature, Linear straight); Cosine is the mirror image, decaying monotonically from $\gamma_{\max}$ at $t=0$ to $\gamma_{\min}$ at $t=T$.
 
\subsection{KL Annealing Schedules $\beta(t)$}
\label{app:beta}
 
Let $T_{\mathrm{VAE}}$ denote the total number of training epochs of the $\beta$-VAE encoder $\varphi$. We define five functional families for the KL weight $\beta(t)$, each parameterized by $(\beta_{\min}, \beta_{\max})$:
\begin{align}
\textbf{Constant:} \quad & \beta(t) = \beta_{\max} \\
\textbf{Linear:} \quad & \beta(t) = \beta_{\max} \cdot \min\!\left(1, \tfrac{t}{T_{\mathrm{VAE}}}\right) \\
\textbf{Cosine:} \quad & \beta(t) = \beta_{\min} + \tfrac{1}{2} (\beta_{\max} - \beta_{\min}) \left(1 + \cos\!\left(\pi \tfrac{t}{T_{\mathrm{VAE}}}\right)\right) \\
\textbf{Sine:} \quad & \beta(t) = \beta_{\max} - \tfrac{1}{2} (\beta_{\max} - \beta_{\min}) \left(1 + \cos\!\left(\pi \tfrac{t}{T_{\mathrm{VAE}}}\right)\right) \\
\textbf{Cyclical:} \quad & \beta(t) = \beta_{\max} \cdot \frac{t \bmod T_{\mathrm{cyc}}}{T_{\mathrm{cyc}}}
\end{align}
 
Linear and Sine implement standard KL warmup (low $\beta$ early, $\beta_{\max}$ at convergence); Cosine inverts the profile, decaying from $\beta_{\max}$ at initialization to $\beta_{\min}$. The Cyclical schedule alternates between regularized and reconstruction-emphasized regimes to mitigate posterior collapse on high-cardinality categorical features.
 
\subsection{Reference Batching}
\label{app:batching}
 
The reference batch $x_r$ used in Equation~\ref{eq:score} is sampled at the start of each generation and held fixed across all reverse-diffusion steps. Two batching strategies are exposed to the TPE search:
 
\textbf{Class-conditional batching.} For classification tasks, $x_r$ is drawn from the same target class as the candidate batch $x_t$. This concentrates the gradient $\nabla_{x_t} \mathcal{C}$ on the conditional manifold relevant to the candidate's target class.
 
\textbf{Global batching.} $x_r$ is drawn uniformly from $D_{\mathrm{real}}$. It is the default for regression, where there is no class structure, and is exposed as an alternative for classification tasks where class-conditional sampling under-represents the global manifold structure.
 
Per-dataset selection is reported in Appendix~\ref{app:configurations}, Table~\ref{tab:configs}.
 
\section{Stage III Sampler Definitions}
\label{app:samplers}
 
This appendix defines the five sampler variants composing the Stage~III grid: the Chamfer Sampler (Section~\ref{app:samplers-chamfer}, recap from main text), and the four alternatives Stratified, IBOSS, HDBSCAN, and MDSampler (Sections~\ref{app:samplers-stratified}--\ref{app:samplers-md}). All samplers operate on the candidate pool $D_{\mathrm{cand}}$ of cardinality $M \cdot N_r$ produced by Stage~II and return a subset of cardinality $K \cdot N_r$ that constitutes the synthetic dataset $D_{\mathrm{syn}}$.
 
\subsection{Chamfer Sampler}
\label{app:samplers-chamfer}
 
The Chamfer Sampler (Section~\ref{sec:method-stage3}) partitions $D_{\mathrm{cand}}$ into batches $\{B_k\}$ of size $b$, ranks each by $\mathcal{C}(B_k, D_{\mathrm{real}})$ (Equation~\ref{eq:sampler}), and retains the union of the top-ranked batches up to $K \cdot N_r$ samples. It is the discrete counterpart of Stage~II's score perturbation, both minimizing $\mathcal{C}$ in $\varphi$-space.
 
\subsection{Stratified Sampler}
\label{app:samplers-stratified}
 
The Stratified Sampler partitions $D_{\mathrm{cand}}$ along the target variable for classification tasks, or along quantile bins of the target for regression, and samples uniformly within each stratum to match the per-class frequencies of $D_{\mathrm{real}}$. The selection is independent of feature-space proximity, retaining only the marginal target distribution of the real set.
 
\subsection{IBOSS Sampler}
\label{app:samplers-iboss}
 
IBOSS (Information-Based Optimal Subdata Selection)~\cite{wang2019information} selects candidates that span the feature space, ensuring every region of the candidate manifold has comparable representation in $D_{\mathrm{syn}}$. It favors extremal points at the boundary of the candidate distribution and is target-agnostic, operating independently of $D_{\mathrm{real}}$.
 
\subsection{HDBSCAN Sampler}
\label{app:samplers-hdbscan}
 
The HDBSCAN Sampler clusters $D_{\mathrm{cand}}$ in $\varphi$-space using the HDBSCAN algorithm~\cite{campello2013density} with $\min\!\_\mathrm{cluster}\!\_\mathrm{size} = b$, retaining the top $K \cdot N_r$ clusters by density. It exploits the manifold structure of the candidate pool but is one-directional with respect to the real-data manifold: synthetic-pool clusters are selected without aligning to $D_{\mathrm{real}}$.
 
\subsection{MDSampler}
\label{app:samplers-md}
 
The MDSampler (Manifold-Distance Sampler) approximates the support of $\varphi(D_{\mathrm{real}})$ as the union of $k$-nearest-neighbor \textit{hyperspheres} around each real point in $\varphi$-space, with each \textit{hypersphere} centered at $\varphi(r)$ and radius $\rho_k(r)$ equal to the distance from $\varphi(r)$ to its $k$-th nearest neighbor in $\varphi(D_{\mathrm{real}})$ (we use $k = 10$). Each candidate $x \in D_{\mathrm{cand}}$ is scored by its distance to this estimated manifold,
\begin{equation}
\label{eq:mdsampler}
d_{\mathrm{MD}}(x) = \min_{r \in D_{\mathrm{real}}} \max\!\big(0,\, \|\varphi(x) - \varphi(r)\|_2 - \rho_k(r)\big) \,,
\end{equation}
which is zero whenever $x$ lies inside any \textit{hypersphere} and equals the distance from the nearest \textit{hypersphere} boundary otherwise. The MDSampler retains the $K \cdot N_r$ candidates with smallest $d_{\mathrm{MD}}$, prioritizing candidates inside the real-data manifold and falling back on the closest off-manifold ones once the inside set is exhausted. The construction follows the manifold Precision proxy of~\cite{kynkaanniemi2019improved}: $d_{\mathrm{MD}} = 0$ corresponds to the inside-the-manifold criterion that defines that metric. The MDSampler is the fidelity-driven counterpart of the bidirectional Chamfer Sampler: when Stage~II is disabled, the candidate pool retains broad coverage by construction, the coverage term $m_{\mathrm{rs}}$ of Equation~\ref{eq:sampler} is approximately satisfied a priori, and inside-manifold selection suffices; when Stage~II is enabled and concentrates the pool around $\varphi(D_{\mathrm{real}})$, most candidates already satisfy $d_{\mathrm{MD}} \approx 0$ and the bidirectional Chamfer Sampler is needed to discriminate via coverage.
 
\section{Generative Knowledge Distillation Details}
\label{app:gkd}
 
This appendix details the Generative Knowledge Distillation (GKD) component of Stage~III, introduced in Section~\ref{sec:method-stage3}.
 
\textbf{Procedure.} GKD trains a teacher classifier $f_{\mathrm{GKD}} : \mathcal{X} \to \Delta^{|\mathcal{Y}|-1}$ on $D_{\mathrm{real}}$, where $\Delta^{|\mathcal{Y}|-1}$ denotes the $(|\mathcal{Y}|-1)$-dimensional probability simplex. For each candidate $\hat{x}_j \in D_{\mathrm{cand}}$ produced by Stage~II, the hard target $\hat{y}_j$ is replaced by the soft-label distribution $f_{\mathrm{GKD}}(\hat{x}_j)$, yielding the relabeled set
\begin{equation}
\label{eq:gkd}
D_{\mathrm{cand}}^{\mathrm{GKD}} = \big\{ (\hat{x}_j,\, f_{\mathrm{GKD}}(\hat{x}_j)) \big\}_{j=1}^{|D_{\mathrm{cand}}|} \,,
\end{equation}
which is then passed to Stage~III's sampler in place of $D_{\mathrm{cand}}$. The downstream learner $A$ trains under cross-entropy with soft targets. GKD relabels targets only, leaving feature vectors $\hat{x}_j$ unchanged.
 
\textbf{Teacher classifier.} In our experiments $f_{\mathrm{GKD}}$ is XGBoost trained on $D_{\mathrm{real}}$ under the same hyperparameter search as the downstream learner $A$, with predicted-probability outputs used as soft labels. Matching $f_{\mathrm{GKD}}$ to the architectural family of $A$ reduces teacher-student mismatch.
 
\textbf{Restriction to classification.} GKD is structurally undefined for continuous regression targets: soft labels over $\mathbb{R}$ require a parametric assumption (e.g., Gaussian conditional) unrelated to the BCR mechanism. GKD is therefore disabled by construction on the 6 regression datasets, halving the Stage~III grid on those.
 
\textbf{Selection regime.} The empirical regime in which GKD is selected is characterized in Section~\ref{sec:results-bcr} (Low-cardinality regime).
 
\section{Per-Dataset Configurations}
\label{app:configurations}
 
This appendix reports the per-dataset structural attributes of the 15 corpus benchmarks (Section~\ref{app:configs-summary}, Table~\ref{tab:dataset_summary}), the search space optimized by the Tree-structured Parzen Estimator over Stage~II hyperparameters (Section~\ref{app:configs-tpe}, Table~\ref{tab:tpe-search-space}), and the configuration retained per dataset against the validation split (Section~\ref{app:configs-selected}, Table~\ref{tab:configs}).
 
\subsection{Dataset Summary}
\label{app:configs-summary}
 
Table~\ref{tab:dataset_summary} reports the structural attributes of the 15 corpus benchmarks, sourced from UCI and Kaggle. Three datasets contain missing values (Adult, Beijing, Infrared Thermography), imputed identically across all models: median for numerical features, mode for categorical, per training split.
 
\begin{table}[h]
\centering
\footnotesize
\setlength{\tabcolsep}{4pt}
\renewcommand{\arraystretch}{1.18}
\caption{Dataset summary. $n$ = number of instances; $d$ = number of feature columns excluding the target; $d_n / d_c$ = numerical / categorical feature counts; NaN indicates presence of missing values (imputed identically across all models, see Appendix~\ref{app:configurations}).}
\label{tab:dataset_summary}
\begin{tabular}{@{}llrrrrc@{}}
\toprule
Dataset                & Task   & $n$            & $d$ & $d_n$ & $d_c$ & NaN \\
\midrule
\multicolumn{7}{l}{\textit{Binary classification}} \\
Adult                  & BC     & $48{,}842$     & 14  & 6     & 8     & Yes \\
Bank                   & BC     & $45{,}211$     & 16  & 7     & 9     & No  \\
Default                & BC     & $30{,}000$     & 23  & 14    & 9     & No  \\
Diabetic Retinopathy   & BC     & $1{,}151$      & 19  & 19    & 0     & No  \\
Magic                  & BC     & $19{,}020$     & 10  & 10    & 0     & No  \\
\midrule
\multicolumn{7}{l}{\textit{Multiclass classification}} \\
Contraceptive          & MC (3) & $1{,}473$      & 9   & 2     & 7     & No  \\
Diamonds               & MC (5) & $53{,}940$     & 9   & 6     & 3     & No  \\
Music                  & MC (11)& $25{,}709$     & 14  & 13    & 1     & No  \\
Student Performance    & MC (4) & $666$          & 11  & 0     & 11    & No  \\
\midrule
\multicolumn{7}{l}{\textit{Regression}} \\
Abalone                & R      & $4{,}177$      & 8   & 7     & 1     & No  \\
Beijing                & R      & $43{,}824$     & 11  & 6     & 5     & Yes \\
Infrared Thermography  & R      & $1{,}020$      & 33  & 30    & 3     & Yes \\
Insurance              & R      & $987$          & 10  & 3     & 7     & No  \\
News                   & R      & $39{,}644$     & 46  & 44    & 2     & No  \\
Solar Flare            & R      & $1{,}066$      & 10  & 7     & 3     & No  \\
\bottomrule
\end{tabular}
\end{table}
 
\subsection{TPE Search Space}
\label{app:configs-tpe}
 
Table~\ref{tab:tpe-search-space} reports the Stage~II hyperparameter search space optimized per dataset by the Tree-structured Parzen Estimator. The $\beta$-VAE inner study (latent dimension and KL annealing schedule) is invoked only when the outer search selects $\beta$-VAE mapping; its functional families are reported in Appendix~\ref{app:beta}.
 
\begin{table}[h]
\centering
\footnotesize
\setlength{\tabcolsep}{4.5pt}
\renewcommand{\arraystretch}{1.2}
\caption{Stage~II hyperparameter search space optimized per dataset by the Tree-structured Parzen Estimator. Domains follow Optuna conventions: \textit{Cat} = categorical, \textit{Int} = integer-valued with the indicated step, \textit{Log} = log-uniform, \textit{Lin} = linear with the indicated step. The $\beta$-VAE inner study (latent dimension and KL annealing schedule) is invoked only when the outer search selects $\beta$-VAE mapping; its search space is reported in Appendix~\ref{app:schedules}.}
\label{tab:tpe-search-space}
\begin{tabular}{@{}llll@{}}
\toprule
Parameter             & Domain                                       & Description                          & Scale     \\
\midrule
$K$                   & $[0.1, 25.0]$                                & Final cardinality multiplier         & Lin (0.1) \\
Guidance variant      & \{None, identity, $\beta$-VAE\}              & Toggle and form of $\varphi$         & Cat       \\
Schedule $\Gamma(t)$  & \{constant, linear, cosine, sine\}           & Guidance scaling family              & Cat       \\
$\gamma_{\max}$       & $[10^{-1}, 5.0]$                             & Upper bound of $\Gamma(t)$           & Log       \\
$\gamma_{\min}$       & $[10^{-8}, \gamma_{\max} - 0.1]$             & Lower bound of $\Gamma(t)$           & Log       \\
$t_g$                 & $[3, 20]$                                    & Guidance step count                  & Int (1)   \\
Batching ratio        & \{None, $1$, $1/2$, $1/4$, $1/8$, $1/10$\}   & Class-conditional fraction           & Cat       \\
\bottomrule
\end{tabular}
\end{table}
 
\subsection{Selected Per-Dataset Configurations}
\label{app:configs-selected}
 
Table~\ref{tab:configs} reports the configuration retained per dataset by the search procedure against the validation split.
 
\begin{table}[h]
\centering
\footnotesize
\setlength{\tabcolsep}{4pt}
\renewcommand{\arraystretch}{1.2}
\caption{Per-dataset configurations selected by the search procedure against the validation split. Stage~II hyperparameters from TPE optimization; Stage~III combination from grid search over $\{\text{GKD off}, \text{GKD on}\} \times \{\text{Stratified}, \text{IBOSS}, \text{HDBSCAN}, \text{MDSampler}, \text{Chamfer Sampler}\}$. Columns: $K$ cardinality multiplier; $G$ guidance toggle (\xmark{} disabled, \cmark{} enabled); $\Gamma$ schedule family (Section~\ref{app:gamma}); $\varphi$ representation map ($\beta$-VAE($d$) for $\beta$-VAE mapping with latent dimension $d$, identity for identity mapping, dash when guidance is disabled); $B_{\mathrm{ratio}}$ reference batching fraction; Type class-conditional (\textit{cls}) or global (\textit{glob}). CS = Chamfer Sampler, MD = MDSampler.}
\label{tab:configs}
\begin{tabular}{@{}lcccccccc@{}}
\toprule
& \multicolumn{6}{c}{Stage~II (TPE)} & \multicolumn{2}{c}{Stage~III (grid)} \\
\cmidrule(lr){2-7}\cmidrule(l){8-9}
Dataset              & $K$  & G       & $\Gamma$  & $\varphi$              & $B_{\mathrm{ratio}}$ & Type & GKD     & Sampler \\
\midrule
Adult                & 0.5  & \color{pred}\xmark & --        & --                     & --      & cls  & \color{pgreen}\cmark & CS      \\
Bank                 & 24.9 & \color{pred}\xmark & --        & --                     & --      & cls  & \color{pred}\xmark & MD      \\
Default              & 13.1 & \color{pgreen}\cmark & cosine    & identity                     & 1/4     & cls  & \color{pred}\xmark & CS      \\
Diabetic Retin.      & 22.5 & \color{pgreen}\cmark & cosine    & $\beta$-VAE(4)  & 1/2     & cls  & \color{pred}\xmark & CS      \\
Magic                & 23.9 & \color{pred}\xmark & --        & --                     & --      & cls  & \color{pred}\xmark & MD      \\
\midrule
Contraceptive        & 8.5  & \color{pgreen}\cmark & sine      & $\beta$-VAE(16) & 1/4     & cls  & \color{pred}\xmark & CS      \\
Diamonds             & 23.1 & \color{pgreen}\cmark & sine      & $\beta$-VAE(4)  & 1/4     & cls  & \color{pred}\xmark & CS      \\
Music                & 20.0 & \color{pred}\xmark & --        & --                     & --      & cls  & \color{pred}\xmark & MD      \\
Student Perf.        & 15.2 & \color{pgreen}\cmark & cosine    & $\beta$-VAE(16) & 1/8     & cls  & \color{pred}\xmark & CS      \\
\midrule
Abalone              & 23.0 & \color{pred}\xmark & --        & --                     & --      & glob & \color{pred}\xmark & MD      \\
Beijing              & 20.0 & \color{pgreen}\cmark & cosine    & identity                     & 1       & glob & \color{pred}\xmark & CS      \\
Infrared Therm.      & 12.8 & \color{pgreen}\cmark & cosine    & $\beta$-VAE(32) & 1/2     & glob & \color{pred}\xmark & CS      \\
Insurance            & 10.0 & \color{pgreen}\cmark & sine      & identity                     & 1/10    & glob & \color{pred}\xmark & CS      \\
News                 & 14.8 & \color{pgreen}\cmark & constant  & $\beta$-VAE(4)  & 1/8     & glob & \color{pred}\xmark & CS      \\
Solar Flare          & 0.1  & \color{pred}\xmark & --        & --                     & --      & glob & \color{pgreen}\cmark & CS      \\
\bottomrule
\end{tabular}
\end{table}
 
\textbf{Aggregate selection counts.} The configuration space partitions cleanly along the Stage~II guidance toggle. Of the 9 datasets enabling guidance, 3 select identity mapping in the data space (Default, Beijing, Insurance) and 6 select $\beta$-VAE mapping in the encoder's latent space (Diabetic Retinopathy, Contraceptive, Diamonds, Student Performance, Infrared Thermography, News); all 9 select the Chamfer Sampler in Stage~III. Of the 6 datasets disabling guidance, 4 select the MDSampler (Bank, Magic, Music, Abalone) and 2 select the Chamfer Sampler with GKD enabled (Adult $K{=}0.5$, Solar Flare $K{=}0.1$, both within the low-cardinality regime $K \leq 1$). GKD is therefore selected on 2 of 15 datasets, both classification datasets at low $K$. Reference batching is class-conditional on all 9 classification datasets and global on all 6 regression datasets, consistent with the structural distinction between the two task families.
 
\section{Auxiliary Metrics}
\label{app:auxiliary-metrics}
 
This appendix reports auxiliary metrics complementing the AUROC, weighted-AUROC, and RMSE numbers of Table~\ref{tab:results}: F1 for classification benchmarks (Section~\ref{app:aux-mle}) and per-dataset DCR-1 share and NNDR breakdowns (Section~\ref{app:aux-additional}). The fidelity (Precision) and diversity (Recall) numbers are reported aggregated in Table~\ref{tab:fidelity_privacy} of the main text.
 
\subsection{Classification F1}
\label{app:aux-mle}
 
\begin{table}[h]
\centering
\footnotesize
\setlength{\tabcolsep}{4pt}
\renewcommand{\arraystretch}{1.18}
\caption{Per-dataset auxiliary downstream classification metrics under TSTR: F1 for binary classification, macro-F1 for multiclass classification. Best per row in bold.}
\label{tab:aux-mle}
\begin{tabular}{@{}lcccccc@{}}
\toprule
Dataset              & Metric  & Real   & TabDiff & TARDIS$\times$1 & TARDIS \\
\midrule
Diabetic Retin.      & F1      & 0.5329 & 0.5469  & 0.5999          & \textbf{0.7407} \\
Adult                & F1      & \textbf{0.707} & 0.671  & 0.696    & 0.704           \\
Default              & F1      & 0.447  & 0.443   & 0.460           & \textbf{0.460}  \\
Magic                & F1      & 0.862  & 0.817   & 0.837           & \textbf{0.880}  \\
Bank                 & F1      & 0.538  & 0.463   & 0.503           & \textbf{0.562}  \\
\midrule
Student Perf.        & macro-F1 & 0.5076 & 0.5119  & 0.3972         & \textbf{0.7299} \\
Contraceptive        & macro-F1 & 0.4229 & 0.4622  & 0.4671         & \textbf{0.5322} \\
Music                & macro-F1 & 0.364  & 0.470   & 0.471          & \textbf{0.489}  \\
Diamonds             & macro-F1 & 0.801  & 0.774   & 0.787          & \textbf{0.804}  \\
\bottomrule
\end{tabular}
\end{table}
 
\textbf{F1 dominance.} TARDIS achieves the best F1 on 8 of 9 classification datasets, with the only exception on Adult ($0.707$ Real vs.\ $0.704$ TARDIS). Gains concentrate on datasets with strong class imbalance or low-cardinality minority classes (Diabetic Retinopathy, Student Performance, Bank), consistent with BCR retaining samples aligned with real-data class-conditional manifolds.
 
\subsection{Per-Dataset Privacy Breakdown}
\label{app:aux-additional}
 
\begin{table}[h]
\centering
\footnotesize
\setlength{\tabcolsep}{4pt}
\renewcommand{\arraystretch}{1.18}
\caption{Per-dataset DCR-1 share (closer to $0.50$ indicates better balance between train-data proximity and novelty) and NNDR (higher indicates more diverse synthetic samples).}
\label{tab:privacy-breakdown}
\begin{tabular}{@{}lcccc@{}}
\toprule
& \multicolumn{2}{c}{DCR-1 share} & \multicolumn{2}{c}{NNDR} \\
\cmidrule(lr){2-3}\cmidrule(l){4-5}
Dataset              & TabDiff & TARDIS  & TabDiff & TARDIS  \\
\midrule
Solar Flare          & 0.4906  & 0.5063  & 0.8876  & 0.8673  \\
Infrared Therm.      & 0.5283  & 0.4902  & 0.9517  & 0.9540  \\
Abalone              & 0.490   & 0.494   & 0.853   & 0.852   \\
Insurance            & 0.487   & 0.487   & 0.762   & 0.773   \\
Beijing              & 0.501   & 0.500   & 0.857   & 0.857   \\
News                 & 0.503   & 0.505   & 0.942   & 0.943   \\
Diabetic Retin.      & 0.5068  & 0.5382  & 0.8706  & 0.8689  \\
Adult                & 0.502   & 0.498   & 0.769   & 0.768   \\
Default              & 0.505   & 0.500   & 0.822   & 0.823   \\
Magic                & 0.501   & 0.491   & 0.882   & 0.879   \\
Bank                 & 0.489   & 0.497   & 0.911   & 0.910   \\
Student Perf.        & 0.5008  & 0.4758  & 0.4970  & 0.5373  \\
Contraceptive        & 0.5061  & 0.5023  & 0.6995  & 0.6871  \\
Music                & 0.506   & 0.504   & 0.883   & 0.877   \\
Diamonds             & 0.500   & 0.499   & 0.761   & 0.761   \\
\bottomrule
\end{tabular}
\end{table}
 
\textbf{Privacy preservation.} TARDIS DCR-1 share remains within $\pm 0.04$ of TabDiff on all 15 datasets, with most absolute deviations under $\pm 0.02$. The three largest deviations (Infrared Thermography $-0.0381$, Diabetic Retinopathy $+0.0314$, Student Performance $-0.0250$) correspond to small-corpus datasets where DCR-1 is more sensitive to the specific subset retained by Stage~III. NNDR is similarly preserved; the largest deviation on Student Performance ($+0.0403$) reflects an unusually low TabDiff baseline ($0.4970$) on a 666-instance dataset where the metric is unstable.
 
\section{Runtime Breakdown}
\label{app:runtime}
 
\begin{figure}[h]
\centering
\includegraphics[width=0.7\linewidth]{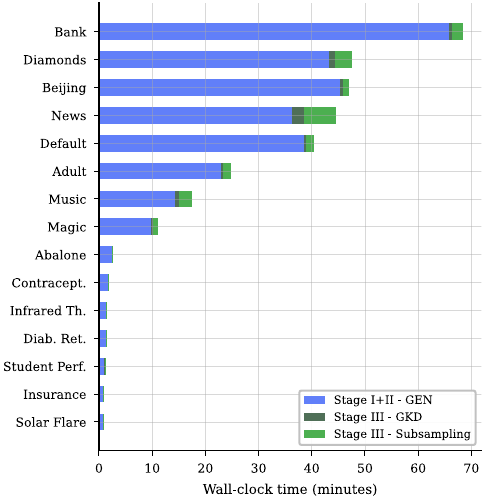}
\caption{Stacked-bar visualization of Table~\ref{tab:runtime}, with stages ordered Stage~II (guided generation), Stage~III (selection), GKD. Datasets sorted by total runtime ascending.}
\label{fig:runtime}
\end{figure}

This appendix details the per-dataset wall-clock time of the TARDIS refinement pipeline (Nvidia GTX 1080~Ti, 12~GB, 2017 consumer GPU). Per-stage runtimes are reported in Table~\ref{tab:runtime} and visualized in Figure~\ref{fig:runtime}. Total times include only the TARDIS refinement passes; TabDiff backbone training is performed once per dataset and reused across configurations.
 
\begin{table}[h]
\centering
\footnotesize
\setlength{\tabcolsep}{3pt}
\renewcommand{\arraystretch}{1.05}
\caption{Per-dataset TARDIS runtime breakdown. Stage~II is the guided generation of $D_{\mathrm{cand}}$ ($M = 50$); Stage~III is the post-hoc selection (Chamfer Sampler or MDSampler); GKD is the soft-label distillation overhead, applicable only when the configuration enables it. Datasets sorted by total runtime ascending.}
\label{tab:runtime}
\begin{tabular}{@{}lrrrr@{}}
\toprule
Dataset              & Stage~II & Stage~III & GKD   & Total \\
                     & (min)    & (min)     & (min) & (min) \\
\midrule
Solar Flare          & 0.7      & 0.1       & 0.1   & 0.9   \\
Insurance            & 0.8      & 0.2       & --    & 1.1   \\
Student Perf.        & 1.0      & 0.2       & --    & 1.4   \\
Diabetic Retin.      & 1.2      & 0.3       & --    & 1.7   \\
Contraceptive        & 1.8      & 0.1       & --    & 2.0   \\
Infrared Therm.      & 1.3      & 0.6       & --    & 2.1   \\
Abalone              & 2.5      & 1.0       & --    & 3.8   \\
Magic                & 9.8      & 4.9       & --    & 16.1  \\
Music                & 14.3     & 11.4      & --    & 29.8  \\
Adult                & 22.9     & 7.0       & 2.3   & 32.2  \\
Default              & 38.5     & 7.1       & --    & 47.9  \\
Beijing              & 45.3     & 4.9       & --    & 53.7  \\
Diamonds             & 43.3     & 16.0      & --    & 65.3  \\
News                 & 36.3     & 29.8      & --    & 77.7  \\
Bank                 & 65.7     & 10.1      & --    & 79.4  \\
\bottomrule
\end{tabular}
\end{table}
 
\end{document}